\documentclass[conference]{IEEEtran}
\usepackage{arydshln}
\IEEEoverridecommandlockouts
\usepackage[font=small,labelfont=bf]{caption}

\usepackage{mwe}
\usepackage{graphicx} % For including images
\usepackage{adjustbox} % To adjust the size of images
\usepackage{booktabs} % To adjust the size of images
\usepackage{lipsum}  
\usepackage{tikz}
\usepackage{multirow}

\usepackage{cite}
\usepackage{amsmath,amssymb,amsfonts}
\usepackage{algorithmic} 
\usepackage{textcomp}
\usepackage{xcolor}
\usepackage{tikz} 
\usepackage{hyperref} 

\hypersetup{
    colorlinks=true,
    linkcolor=blue,
    filecolor=magenta,      
    urlcolor=magenta,
    citecolor=blue
}

\def\BibTeX{{\rm B\kern-.05em{\sc i\kern-.025em b}\kern-.08em
    T\kern-.1667em\lower.7ex\hbox{E}\kern-.125emX}}
\begin{document}
 
\title{

    % Watermark at the top, split into two lines and adjusted
    \begin{tikzpicture}[overlay, remember picture]
        \node at ([yshift=-1.3cm]current page.north) {
            \normalsize\textcolor{gray}{This paper has been accepted for publication at the }
        };
        \node at ([yshift=-1.8cm]current page.north) {
            \normalsize\textcolor{gray}{IEEE International Joint Conference on Neural Networks (IJCNN), Rome, Italy 2025}
        };
    \end{tikzpicture}
    
    % Title
    RGB-Event Fusion with Self-Attention for Collision Prediction 
    
    \thanks{This work was funded by the Swiss National Science Foundation (Grant 219943) and by the SONY "Research Award Program (RAP)".} 
    
    % % Side watermark for arXiv, positioned properly
    % \begin{tikzpicture}[overlay, remember picture]
    %     \node at ([xshift=1.2cm]current page.west |- current page.center) [rotate=90] {
    %         \LARGE\textcolor{gray}{arXiv:2504.10400 [cs.CV] \today}
    %     };
    % \end{tikzpicture}
}

\author{
    \IEEEauthorblockN{Pietro Bonazzi\IEEEauthorrefmark{1}, Christian Vogt\IEEEauthorrefmark{1}, Michael Jost\IEEEauthorrefmark{1},
    Haotong Qin\IEEEauthorrefmark{1}, \\ Lyes Khacef\IEEEauthorrefmark{2}, Federico Paredes-Valles\IEEEauthorrefmark{2}, Michele Magno\IEEEauthorrefmark{1}}
    
    \IEEEauthorblockA{\IEEEauthorrefmark{1}ETH Zürich, Zürich, Switzerland}
    \IEEEauthorblockA{\IEEEauthorrefmark{2}Sony Semiconductor Solutions Europe, Sony Europe B.V., Zürich, Switzerland} 
}

\maketitle

\begin{abstract}

Ensuring robust and real-time obstacle avoidance is critical for the safe operation of autonomous robots in dynamic, real-world environments. 
This paper proposes a neural network framework for predicting the time and collision position of an unmanned aerial vehicle with a dynamic object, using RGB and event-based vision sensors.
The proposed architecture consists of two separate encoder branches, one for each modality, followed by fusion by self-attention to improve prediction accuracy.

To facilitate benchmarking, we leverage the ABCD \cite{Bonazzi2025CVPRW} dataset collected that enables detailed comparisons of single-modality and fusion-based approaches.

At the same prediction throughput of 50Hz, the experimental results show that the fusion-based model offers an improvement in prediction accuracy over single-modality approaches of 1\% on average and 10\% for distances beyond 0.5m, but comes at the cost of +71\% in memory and + 105\% in FLOPs. Notably, the event-based model outperforms the RGB model by 4\% for position and 26\% for time error at a similar computational cost, making it a competitive alternative.

Additionally, we evaluate quantized versions of the event-based models, applying 1- to 8-bit quantization to assess the trade-offs between predictive performance and computational efficiency.

These findings highlight the trade-offs of multi-modal perception using RGB and event-based cameras in robotic applications.

\end{abstract}

\begin{IEEEkeywords}
Drone, TinyML, Obstacle Avoidance, Event-Based Camera
\end{IEEEkeywords}

\section*{Reproducibility}

The code base and dataset can be installed using the link : \url{https://github.com/pbonazzi/eva}.

\section{Introduction}

Unmanned aerial vehicles (UAVs) have revolutionized various fields, including logistics, surveillance, agriculture, and disaster management. As UAVs are deployed in dynamic environments \cite{wang2024dynamic}, efficient and accurate obstacle avoidance is critical for ensuring safety and operational effectiveness.

Traditional frame-based cameras have been widely used for obstacle avoidance due to their high spatial resolution and compatibility with conventional computer vision algorithms\cite{Xu_2023}. However, their limitations such as motion blur, high latency, and low dynamic range hinder their performance in high-speed or low-light scenarios \cite{yasin2020}. In contrast, event cameras, also referred to as event-based vision sensors (EVS), offer a promising alternative by capturing changes in pixel intensity asynchronously, resulting in significantly lower latency, higher temporal resolution, and a wider dynamic range \cite{Gallego_2022, gruel2023simultaneous, Bonazzi_2024_CVPR, Forrai_23_ICRA, monforte2023}. These advantages make event cameras particularly well-suited for applications that require fast and precise motion analysis, including obstacle avoidance \cite{sanket2020evdodgenet, falanga2020, bisulco2021, paredes2024}.

The maturity of the field has led to real-world demonstrators \cite{doi:10.1126/scirobotics.adj8812, 10164354}, underscoring the demand for robust, real-time collision avoidance mechanisms in drone navigation. However, in-field demonstrations introduce significant challenges \cite{loquercio2021}, such as environment noise, and introduce the need for low-latency processing in dynamic environments \cite{tomy2022fusing}. These challenges necessitate multi-sensor fusion strategies.

Event-based cameras capture pixel-level brightness changes with microsecond latency, enabling detection of fast-moving objects even in changing or low-light environments. In contrast, RGB cameras provide rich spatial and color information. By fusing these complementary data streams, researchers have shown it is possible to enhance the robustness of deep learning vision algorithms \cite{10161563, 10508494, 10818066, 10801307, 9359329}.

\begin{figure*}[t!]
    \centering
    \begin{tabular}{c} 
        \includegraphics[width=\textwidth]{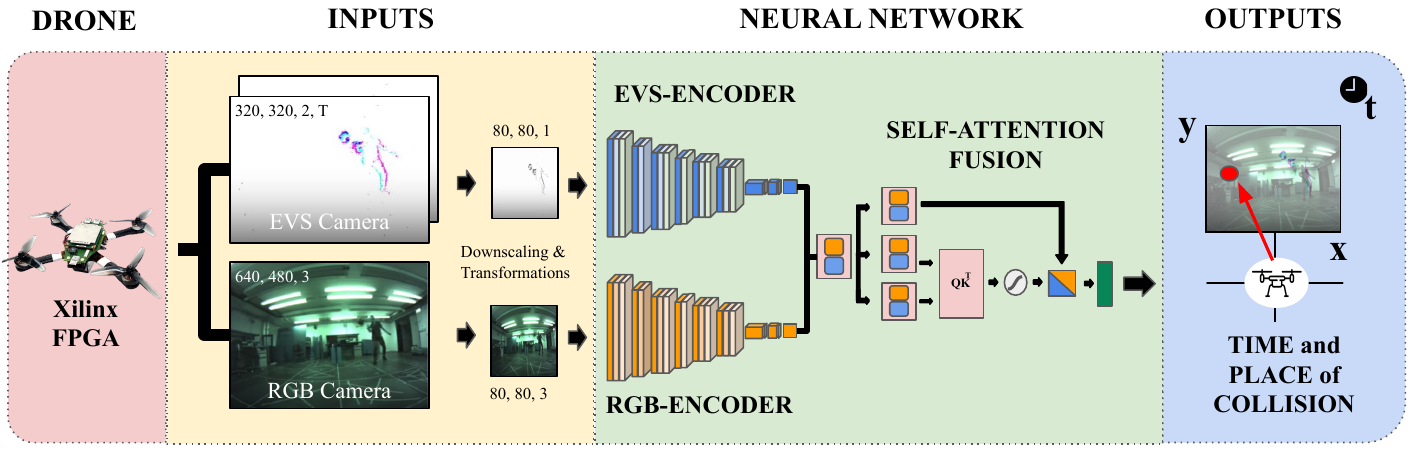} \\ 
    \end{tabular}
    \caption{\textbf{A visual overview of the proposed architecture}: (1) The FPGA-based drone setup; (2) Input data streams, including event-based and frame-based data; (3) Preprocessing and data transformation pipeline; (4) Neural network architectures; (5) Outputs predicting collision time and position.}
    \label{fig:architecture}
\end{figure*}

This work specifically tackles the problem of accurately predicting both the position and time of potential collisions of an object by integrating event-based and RGB data. A visual overview of the proposed framework is provided in Fig.~\ref{fig:architecture}.

\section{Related Work} 

Event-based obstacle avoidance has gained significant attention in recent years due to the advantages offered by event cameras, such as low latency, high temporal resolution, and high dynamic range. These characteristics make event cameras particularly suitable for dynamic obstacle avoidance scenarios.

\subsection{Event-Based Obstacle Avoidance}

\subsubsection{Optical-Flow  Approaches}

Early work on event-based obstacle avoidance focused on optical flow. Clady et al. \cite{Clady2014} proposed an asynchronous time-to-contact algorithm for collision detection based on event-based motion flows. Milde et al. \cite{Milde2015} developed an event-driven collision avoidance algorithm based on optical flow. Yasin et al. \cite{yasin2020} explored the detection and avoidance of obstacles in low light conditions by triangulating 2D features extracted with LC-Harris. Falanga et al. \cite{falanga2020} demonstrated a dynamic obstacle avoidance system for quadrotors using event cameras. More recently, Rodriguez-Gomez et al. \cite{rodriguez2022} demonstrated a dynamic sense-and-avoid system for ornithopter robot flight using optical flow and clustering. 

\subsubsection{Data-Driven Approaches}

Deep learning approaches have also been applied to event-based obstacle avoidance. Sanket et al. \cite{sanket2020evdodgenet} and Bhattacharya et al. \cite{bhattacharya2024monocular} developed a static-obstacle avoidance method for quadrotors using a monocular event camera.. Bisulco et al. \cite{bisulco2021} proposed a method to understand fast motion using spatio-temporal neural networks of a moving toy dart. Paredes-Valles et al. \cite{paredes2024} trained a spiking neural network that accepts raw event-based camera data and outputs low-level control actions to perform autonomous vision-based flight. Differently from these approaches, Ours is multi-modal, lightweight and evaluated on an open-source dataset, namely the Ball-Collision Dataset, which can serve as a benchmark for further data-driven algorithmic exploration.

\subsection{Event-Based Object Catching}

Related to obstacle avoidance, Forrai et al. \cite{Forrai_23_ICRA} showcased agile object catching based on events with a quadrupedal robot. Monforte et al. \cite{monforte2023} proposed the use of event cameras for trajectory prediction with a robotic arm, leveraging a stateful LSTM network trained in simulation. Notably, the event-based method achieved higher accuracy compared to the respective traditional frame-based version. Our work builds upon this evaluation framework while also integrating event-based perception with a more sophisticated multi-modal fusion approach and leveraging lightweight stateless networks.

\subsection{Multi-Modal Fusion Approaches with Event Cameras}

Walters et al. \cite{walters2021}  and He et al. \cite{he2021} developed a system for the detection and tracking of dynamic objects using both event and depth sensing. Devulapally et al. \cite{Devulapally_2024_CVPR} proposed a unified transformer-based architecture that fuses event and RGB data to improve monocular depth estimation accuracy. Zhou et al. \cite{10161563} introduced RENet, an RGB-event fusion network that enhances moving object detection in autonomous driving using temporal and bi-directional fusion modules. Cre\ss{} et al. \cite{10508494} developed the TUMTraf Event dataset with synchronized event and RGB data, offering various fusion strategies to improve detection in transportation systems. Fradi and Papadakis \cite{10818066} presented an event-RGB fusion framework that probabilistically combines detections from both modalities to enhance object detection in autonomous vehicles. These studies highlight the potential of event-RGB fusion in enhancing perception tasks. Our methodology adopts similar strategies, fusing event and RGB data to improve prediction.

\section{METHODOLOGY}

\subsection{Dataset}

This study leverages the ABCD dataset, originally introduced in our prior work \cite{Bonazzi2025CVPRW}, which was built on the SwiftEagle platform \cite{vogt2024IROS}—an open-source, FPGA-based UAV designed for low-latency sensing and real-time decision-making. The SwiftEagle system integrates a frame-based RGB camera (SONY IMX219) and an event-based sensor (Prophesee GenX320), connected via a custom FPGA pipeline capable of handling dual-camera input for efficient multi-modal processing.

The dataset was collected using a lightweight foam ball (15 cm, 65 g) embedded with IR LEDs, tracked in 3D space via a VICON motion capture system. The ball was thrown toward a drone protected by a plexiglass barrier, which introduced real-world challenges such as reflection and vibration artifacts in the visual data. The dataset comprises 206 multi-modal recordings of ball throws, with experiments focusing on the last 30 seconds of each sequence, corresponding to the approach and impact phase of the trajectory.

The ABCD dataset includes diverse motion patterns with elevation angles from 0.017 m to 2.009 m, flight lengths from 0.502 m to 2.469 m, and speeds between 0.329 m/s and 7.739 m/s. These variations support robust evaluation of motion prediction, collision dynamics, and sensor fusion strategies.

All recordings were conducted in a controlled VICON-equipped arena, where both the drone and the ball were fitted with reflective markers for sub-millimeter accurate ground truth collection. Dataset splits (train/val/test1/test2) were constructed by first categorizing recordings based on whether the collision point occurred within a central 600 mm × 600 mm region around the drone, and then ensuring balanced sampling across both spatial regions.

Together, these features establish Ball-Collision as a challenging, real-world dataset suitable for benchmarking multi-modal perception models, and form part of our ongoing efforts to expand the NeuroBench framework \cite{yik2025neurobenchframeworkbenchmarkingneuromorphic}.

\subsection{Input Representation}

The RGB and event representations are designed to facilitate efficient memory management and computational deployment on FPGA hardware.

For the event-based input representation, event streams are processed using a fixed time window of duration $T$ ($T=20$ms unless otherwise noted). Within this window, events are accumulated on a per-pixel basis, separately for both positive and negative polarities. The final event representation is obtained by computing the pixel-wise difference between the two polarities, resulting in a tensor of dimensions $80 \times 80 \times 1$. 

For the RGB input representation, raw frames ($T$=20ms) are captured and resized to a fixed resolution of $80 \times 80 \times 3$. Similar to event data, no temporal multi-dimension is applied to the RGB input unless explicitly stated. While better pre-processing approaches have been proposed \cite{Zubic_2023_ICCV}, this approach is compatible with efficient FPGA implementations \cite{Bonazzi2025CVPRW} and avoids channel-last, channel-first issues related to the Prophesee GenX320 sensor and AMD Xilinx Deep Learning Unit.

\subsection{Neural Network Architecture}

The proposed model incorporates an encoder-decoder structure that processes both event-based and RGB image inputs through separate encoder branches. 

Each encoder branch 
%shown in Fig. \ref{fig:architecture_detail}, 
features 6 convolutional layers ($3\times 3$ kernel size, stride of 1) followed by ReLU \cite{agarap2019deeplearningusingrectified} activation functions, designed to learn spatial hierarchies from the input data progressively. 

The outputs from the EVS and RGB encoder branches are concatenated to form a unified feature representation and then processed by a self-attention mechanism to dynamically weigh the contributions of different features \cite{Devulapally_2024_CVPR}.

Finally, a fully connected layer transforms the output of the self-attention block into a 3-dimensional prediction: the \(x\) and \(y\) coordinates of the collision relative to the drone, and \(t\), the estimated time to collision. In this last layer, no activation function is used.

\subsection{Low-Bit Variants Single Modality Networks}

As for the efficiency requirements of real deployment on edge hardware, we further compress the proposed neural architecture to lower bit-width. Specifically, we present low-bit quantized and 1-bit binarized variants for different accuracy-efficiency balance options.

\subsubsection{Quantized Variant}

For the quantized variant, the weight and activation in neural architectures are quantized to low bit-width, i.e., 2--8 bits. The quantized parameters allow compressed storage usages of the model (e.g., 8$\times$ compression for 4-bit), and the floating-point computation can be replaced by more efficient integer operations.

We adopt the LSQ+~\cite{bhalgat2020lsq} method, which introduces a general asymmetric quantization scheme with learnable scale and offset parameters to accommodate negative activations, created by the required [-1,1] input range required by the FPGA-accellerator. 

\subsubsection{Binary Variants}

We also evaluated binary versions of the proposed event-based model to explore the trade-off between model efficiency and performance. Different from low-bit quantization, binarization not only allows models to have the most extremely compressed 1-bit-width but also enjoys fast XNOR-POPCNT bitwise instructions during inference \cite{agrawal2019}.

In the binary variant, we tested and compared several state-of-the-art binary neural network methods to improve performance at extreme compression rates. Specifically, we applied the DoReFa-Net \cite{zhou2016dorefa} method, which allows stochastically quantizing the parameter gradients to low bit-width values during the backward pass. 
We also tested IRNet \cite{qin2020forwardbackwardinformationretention}, which focuses on retaining information during both forward activations and backward gradients computation, and a modified version of IRNet (named "IRNet*" in the Experiments Section) without standard weight normalization.
Furthermore, we evaluated ReActNet \cite{liu2020reactnetprecisebinaryneural}, which introduces RSign and RPReLU functions to reshape and shift activation distributions alongside a distributional loss to enforce similarity between binarized and real-valued network outputs. 

\subsection{Loss Function and Training Details}

The task involves predicting the relative XY position of the collision point of a ball of fixed size with respect to the drone's location (static across all sequences) and the time to collision. A plexiglass wall is placed in front of the drone to protect it from possible hardware damage. For this purpose, the loss function used is the Mean Squared Error (MSE) for both predictions. The 3D positions of the drone and the ball are obtained from a Vicon motion capture system. The collision point is computed as the minimum distance between the two 3D trajectories (the ball and the drone) over the recording period.

We used a batch size of 16 and an initial learning rate of 1e-4. The training process incorporated stochastic weight averaging, gradient clipping, and early stopping with a patience of 10 epochs to prevent overfitting. Optimization was performed using the Adam optimizer \cite{kingma2017adammethodstochasticoptimization} with a weight decay of 1e-5. Additionally, a learning rate scheduler was employed to reduce the learning rate on a plateau with a patience of 5 epochs and a reduction factor of 0.1. Dropout regularization is employed at each stage to enhance generalization.

These training strategies ensured stable convergence and prevented overfitting. We also evaluated several compression strategies, introduced in the previous subsection, based on binarization and quantization aware training to improve computational efficiency of our algorithms.

%\begin{figure}[h!]
 %   \centering
  %  \begin{tabular}{c}
        %\includegraphics[width=0.5\textwidth]{media/architecture_detail.pdf}  \\
    %\end{tabular}
    %\caption{\textbf{Diagram of the encoding branch architecture}. The EVS and RGB Encoder feature a series of convolution and pooling layers with kernel size 3x3 and 2x2 respectively.}
     %\label{fig:architecture_detail}
%\end{figure}

\section{EXPERIMENTS}

In this section, we evaluate the performance of our proposed models on the relative position estimation and time to collision error using single-mode setups (i.e., EVS and RGB separately) and the fusion setup (i.e., leveraging both modalities). We compare position and time errors with ground truth data provided by the motion capture system and analyze the results under various conditions, such as distance ranges, frame rates, and compression rates.

\subsection{Fusion Model vs Single Modality Models}

\subsubsection{Precision and Computational Complexity}

\begin{figure}[t]
    \centering
    \includegraphics[width=0.5\textwidth]{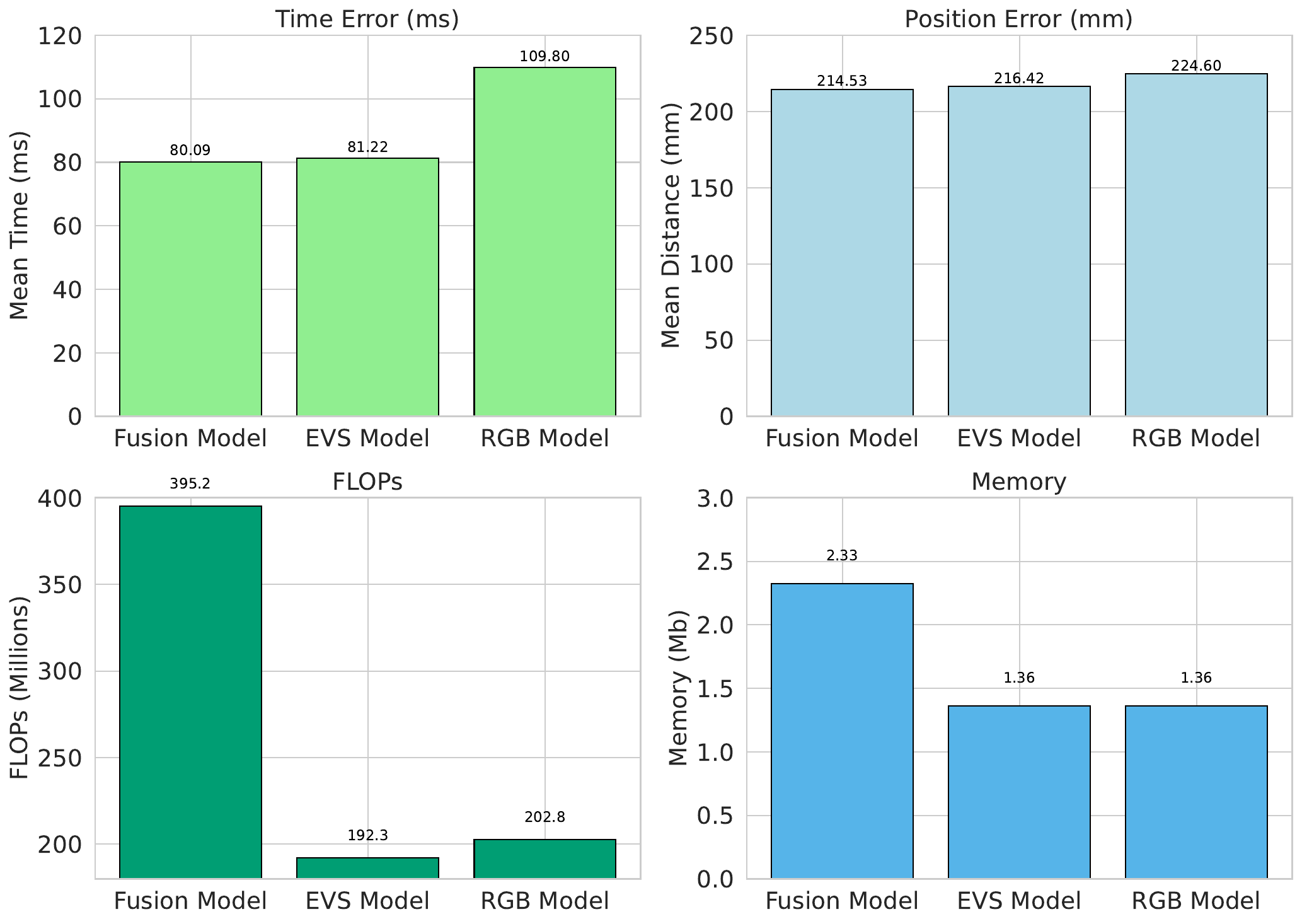}
    \caption{\textbf{Performance comparison of the Fusion, RGB, and EVS models}. The Fusion model demonstrates superior accuracy but requires higher computational resources. FLOPs refers to floating-point operations.}
    \label{fig:multimodal_plots}
\end{figure}

The results in Fig.~\ref{fig:multimodal_plots} demonstrate the effectiveness of the fusion model, which benefits from combining complementary information from both modalities. The fusion model consistently achieves higher precision compared to single-mode setups over all sequences at the cost of computational complexity. 

An error rate of 200mm for the position error and 100ms might in fact be "enough" to execute the drone evasion maneuver in time to avoid collisions. Further real world evaluations, with a flying drone, and analysis of thurst to weight ratio, motor specs are needed to validate this hypothesis.

\subsubsection{Spatial Precision at Different Distances}

\begin{table*}[h] %TABLE 4
	\centering
	% Captions go above tables
	\caption{\textbf{Position error of our models different distance ranges} For each configuration, we report (expressed in millimeters) the mean, the median, the SD, and the maximum absolute deviation (M.A.D.) of the norm of the position error for different ranges of distances.}
	\label{tab:distance_ablation} 
	
    \begin{tabular}{l|cccc|cccc|cccc}
    
     & \multicolumn{4}{c|}{EVS-Model} & \multicolumn{4}{|c|}{RGB-Model} & \multicolumn{4}{|c}{Fusion-Model} \\ 
     
    \cmidrule(lr){2-13}
    
    Range (m) & Mean & Median & SD & M.A.D. & Mean & Median & SD & M.A.D. & Mean & Median & SD & M.A.D. \\ \midrule

    0.2-0.5 & 498.51 & 529.57 & 105.35 & 65.26 & 471.05 & 509.53 & 139.03 & 87.57 & 545.44 & 600.37 & 146.39 & 130.87 
    \\ \midrule 
    
	0.5-1.0  & 232.54 & 136.43 & 207.94 & 179.93 & 225.45 & 146.25 & 188.71 & 160.47  & 228.92 & 140.34 & 210.06 & 174.58  \\ \midrule
    
	1.0-1.5  & 210.27 & 152.68 & 166.43 & 130.02 & 217.14 & 170.10 & 158.45 & 126.67 & 194.54 & 126.10 & 194.28 & 140.63 \\ \midrule
    
	1.5-2.0 & 214.09 & 164.40 & 146.47 & 115.69 & 227.70 & 207.61 & 149.18 & 115.57 & 201.95 & 151.82 & 167.89 & 117.87 \\ \midrule
    
	2.0+ & 193.55 & 166.66 & 106.02 & 84.48 & 215.28 & 209.81 & 124.07 & 92.92 & 202.48 & 167.46 & 112.85 & 86.61 \\ 
	\end{tabular} 
    
\end{table*}

We evaluate the position error of single-modality and fusion-based models at $T=20$ms, quantified as the norm of the deviation from ground truth (Table \ref{tab:distance_ablation}). The analysis covers multiple distance ranges, reporting mean, median, standard deviation (SD), and maximum absolute deviation (M.A.D.).

The fusion model outperforms both single-modality approaches, delivering lower mean and median errors, with the exception of closer range (0.2-0.5m). At close proximity, the fusion model shows a mean error of 545.44mm, which is approximately 9\% higher than the EVS-model (498.51mm) and 15\% higher than the RGB-model (471.05mm). At mid-ranges (1-1.5m), the fusion model shows a mean error of 194mm,, which is 11\% lower than the RGB-model (217mm) and 7\% lower than the EVS-model (210mm). At longer ranges (2.0+ m), the fusion model shows a mean error of 202.48mm, which is 6.4\% lower than the RGB-model (215.28mm), while being 4.3\% higher than the EVS-model (193.55mm).

Single-modality models shows competitive performance and lower variability, making them suitable for dynamic environments requiring low computational complexity. At shorter distances (0.2-0.5m). The EVS-model, shows lower SD (105.35mm) and M.A.D. (65.26mm) compared to RGB-based models (SD: 139.03mm, M.A.D.: 87.57mm).

% \subsubsection{Spatial and Temporal Precision at Different Distances}

\begin{figure}[t!]
    \centering
    \begin{tabular}{c} 
        \includegraphics[width=0.45\textwidth]{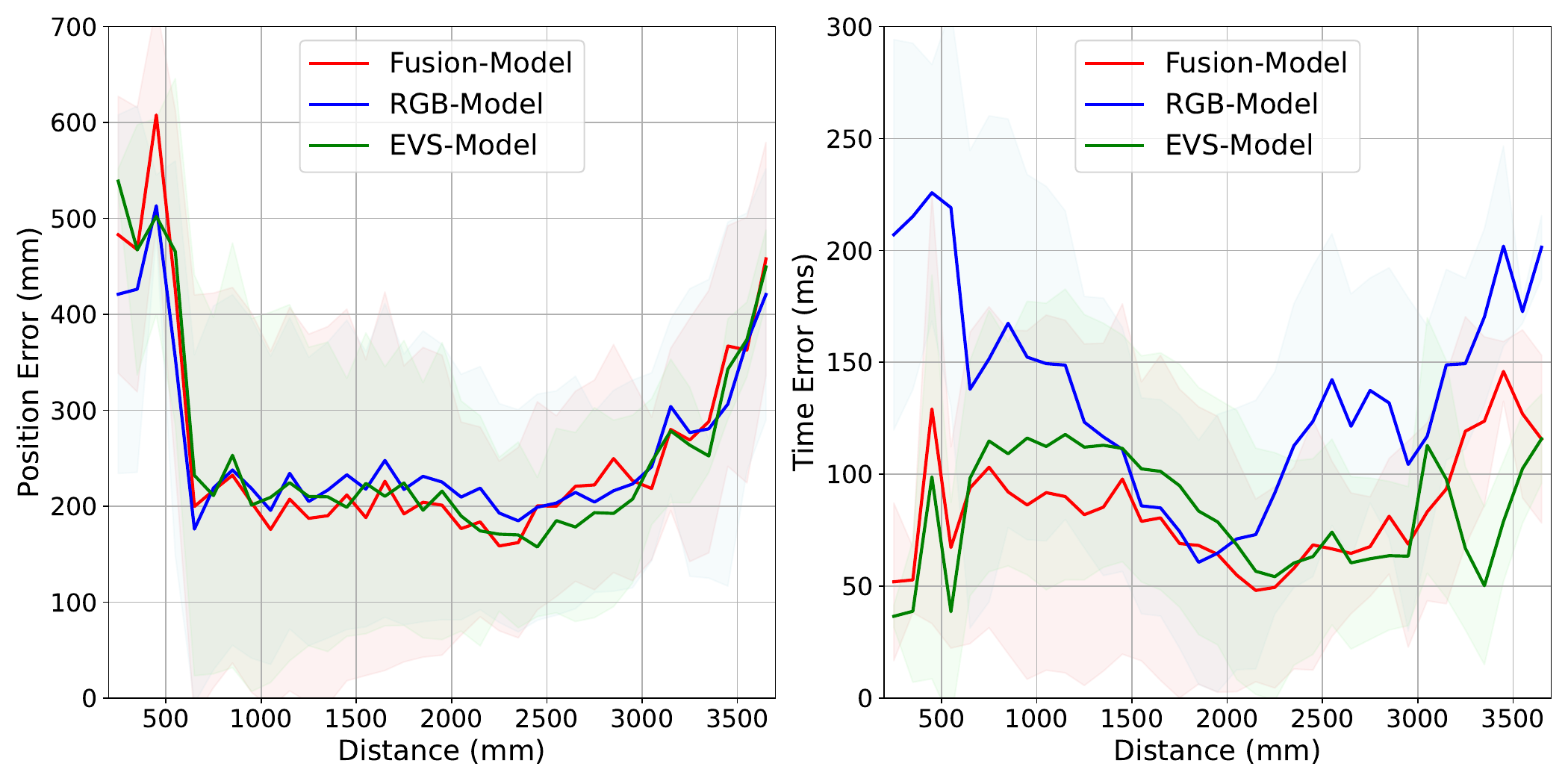} \\ 
    \end{tabular}
    \caption{\textbf{Position and time error analysis of the single-modality and fusion model ($T=dT=20$ms) over true distance of ball w.r.t the collision point.} Results are shown as average and standard deviation over all recordings of the test dataset.}
    \label{fig:test_errors_over_distance_comparison_all}
\end{figure}

% We further investigate the position and time errors of the models at a time interval of $T=20$ms. 
Lastly, Fig. \ref{fig:test_errors_over_distance_comparison_all} shows the position and time error analysis with respect to the true distance of the ball relative to the collision point. The error increases with the distance and at close proximity. Notably, RGB exhibits significantly higher time error, likely due to intrinsic lower temporal resolution, and sensitivity to lighting conditions, making it less reliable compared to the EVS.

\subsection{Single-Modality Ablation Study}

\subsubsection{Precision at different frame rates}

\begin{figure}[t]
    \centering
    \begin{tabular}{c}
        \includegraphics[width=0.46\textwidth]{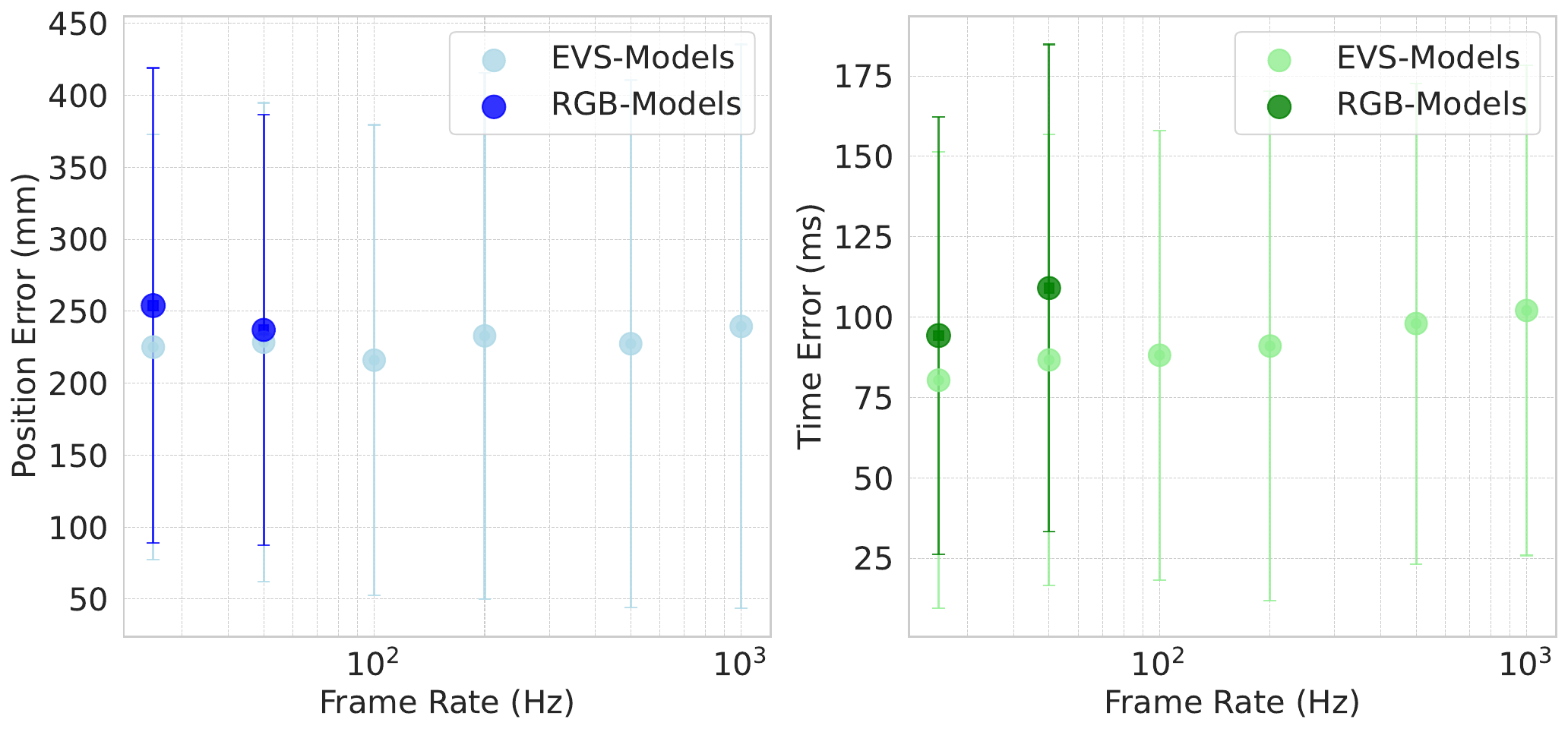}  \\
    \end{tabular}
    \caption{\textbf{Performance of RGB and event-based models at different frame rates.} On average, EVS-only models can increase throughput and outperform RGB models running at similar frame rates.}
    \label{fig:frame_rate_over_error}
\end{figure}

Considering that single-modality models might offer a better trade-off between error and computational cost, we evaluated their performance at different frame rates to assess their suitability for various deployment scenarios. The frame rate for the event-based model is defined as $1/T$. The original frame rate of the RGB camera is fixed at 50 frames per seconds \cite{vogt2024IROS}, thus to expand the comparison we manufactured lower frame rates by concatenating consecutive samples. Concatenating RGB frames would lead to improved performance on the temporal task and a slight degradation in the spatial task, see Fig. \ref{fig:frame_rate_over_error}.
% Considering that the performance of the RGB-model running atr 50 FPS frame at a time and a stateless network, temporal information in the RGB-model comes from motion blur.
A frame rate of 25Hz (corresponding to the first points from the left in both plots, Fig. \ref{fig:frame_rate_over_error}) is achieved by passing two consecutive distinct windows of $T=20$ms on the EVS models. Fig. \ref{fig:frame_rate_over_error} also indicates that EVS-based models consistently outperform RGB-based models across the two frame rates, maintaining lower errors even at higher throughput. In fact, EVS models are capable of operating at higher frame rates than RGB models while still achieving lower position error than the RGB-models running at 50Hz.

Additionally, while lower frame rates improve the time-to-collision prediction for both models by providing larger time windows, they do not alter the performance in position-of-collision estimation significantly, suggesting there might be enough spatial information at $T=1$ms of event data to solve the task. 

\subsubsection{Compression of event-based models}

\begin{figure}[h!]
    \centering
    \begin{tabular}{c}
        \includegraphics[width=0.45\textwidth]{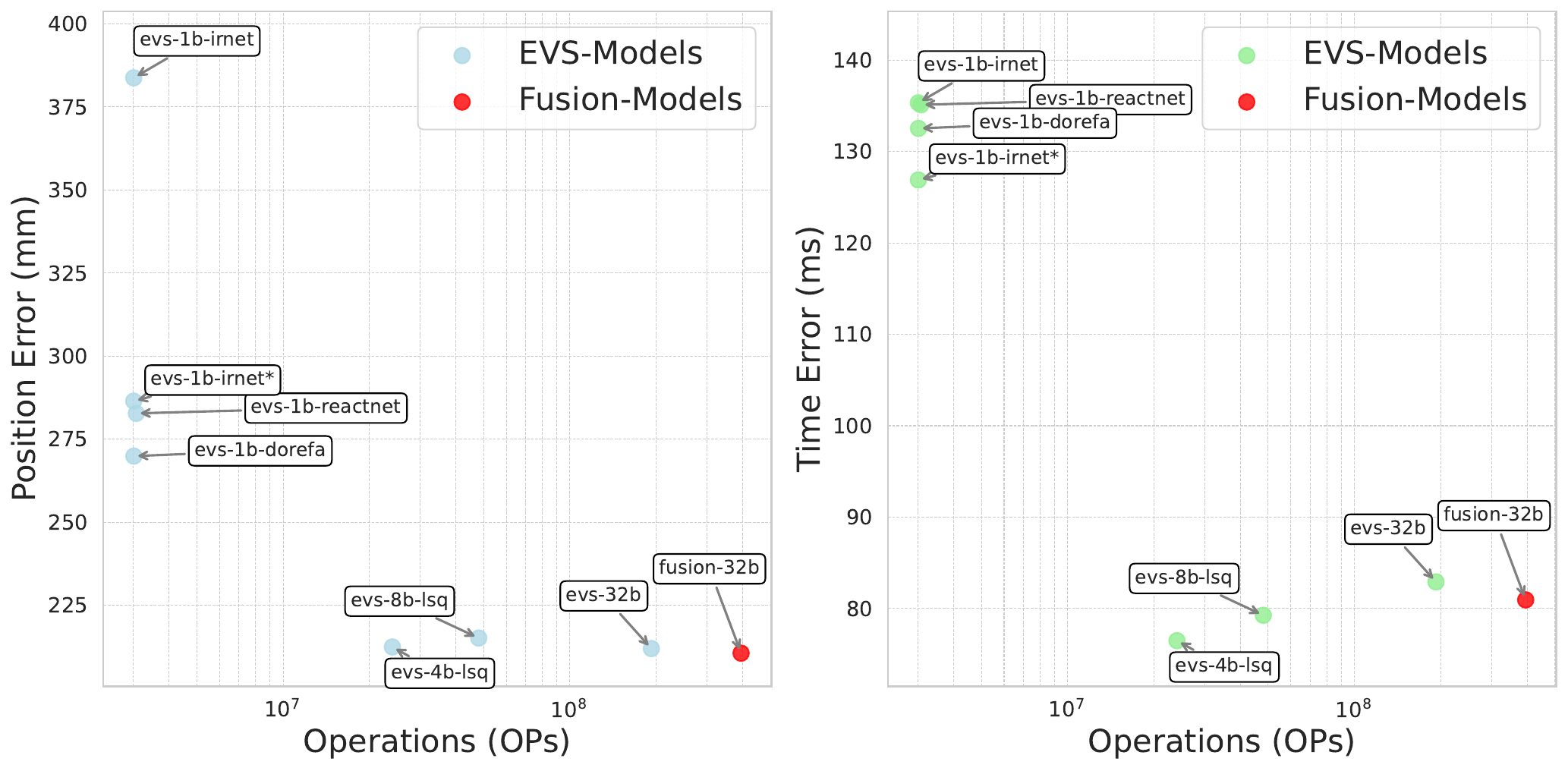}  \\
    \end{tabular}
    \caption{\textbf{Performance Analysis of Event-Based at different compression rates.} The plots show the errors over the total number of operations (OPs).}
    \label{fig:position_time_error_compression}
\end{figure}

Next, we explore the impact of different compression rates on the performance of event-based models. Fig. \ref{fig:position_time_error_compression} presents a performance analysis of the event-based model under different compression rates.

We only notice a significant degradation in the model's performance on the position-to-collision task and in the time-to-collision task on the binarized variants of the models. 

In addition, moving from 32-bit to 8-bit and 4-bit reduces the memory footprint and achieves 4x to 8x efficiency gains compared to floating point 32-bit operations.

Binarization, however, introduces a fundamental shift, replacing multiplication with highly efficient bitwise XNOR \cite{agrawal2019} operations. This drastically reduces the number of operations, often by a factor of 64, due to the inherent parallelization of binary operations in digital hardware \cite{liu2020reactnetprecisebinaryneural}.
In terms of binarization method, no clear winner emerges. However, we note that removing IRNET standard deviation normalization improves results compared to the original method \cite{qin2020forwardbackwardinformationretention}.

Fig. \ref{fig:position_time_error_compression} illustrates this phenomenon by plotting errors against operations (OPs). The total operations are defined as the sum of binary, integer and floating-point operations, calculated as \( \text{OPs} = \frac{\text{BOPs}}{64} + \frac{\text{INT4\_OPs}}{8} + \frac{\text{INT8\_OPs}}{4} + \text{FLOPs} \).

\section{CONCLUSION}

This work proposed a robust framework for predicting the position and time of collision of a ball using event-based and RGB image data. Our contributions include the development of lightweight neural network architectures for single-modality and multi-modality inputs, as well as the evaluation on a novel multi-modal dataset, ABCD \cite{Bonazzi2025CVPRW}.

Experimental results demonstrate that event-based data, when compared to RGB, offers advantages in terms of performance. With the same prediction throughput (50Hz frames from RGB and event frames constructed from EVS), EVS outperforms RGB, reducing position error by approximately 11\% and time error by 25\%. Fusion of EVS and RGB data slightly improves performance over EVS alone, with position error reduced by ~0.5\% and time error reduced by ~2\%. However, this improvement comes at the cost of a 70\% increase in memory usage and a 105\% increase in FLOPs.

These findings suggest that, within the assumptions of this work, EVS presents a better trade-off between accuracy and computational cost. Further studies should explore the effects of higher-frequency EVS for reducing response latency and investigate additional techniques, such as spiking neural networks, to further minimize energy consumption.  Ultimately, this work paves the way toward faster and more efficient perception systems for autonomous drones and robotics systems with on-device AI capabilities.

\bibliographystyle{plain}  
\bibliography{bibliography}

\end{document}